# New Method for Localization and Human Being Detection using UWB Technology: Helpful Solution for Rescue Robots*

David Espès, Ana Maria Pistea, Charles Canaff, Ioana Iordache, Philippe Le Parc, *Member, IEEE*
Emanuel Radoi, *Member, IEEE*

*Abstract*— **Two challenges for rescue robots are to detect human beings and to have an accurate positioning system. In indoor positioning, GPS receivers cannot be used due to the reflections or attenuation caused by obstacles. To detect human beings, sensors such as thermal camera, ultrasonic and microphone can be embedded on the rescue robot. The drawback of these sensors is the detection range. These sensors have to be in close proximity to the victim in order to detect it. UWB technology is then very helpful to ensure precise localization of the rescue robot inside the disaster site and detect human beings.**

**We propose a new method to both detect human beings and locate the rescue robot at the same time. To achieve these goals we optimize the design of UWB pulses based on B-splines. The spectral effectiveness is optimized so the symbols are easier to detect and the mitigation with noise is reduced. Our positioning system performs to locate the rescue robot with an accuracy about 2 centimeters. During some tests we discover that UWB signal characteristics abruptly change after passing through a human body. Our system uses this particular signature to detect human body.**

## I. Introduction

Nowadays most of the people of a country live in urban centers. As cities grow, their population density increases. The high population density makes any disaster, such as natural or human disaster, much more deadly. Another impact of megacities is the complexity to deploy rescue teams [1]. Indeed a disaster may destroy the critical infrastructures of a city and reduces the effectiveness of the rescue teams greatly. Disaster sites are complex and dangerous. They are a great threat to rescue workers and survivors. In such chaotic scenarios, the rescue workers' lives can be put at risk. In order to avoid unnecessary losses, a rescue robot can be of great help to detect human beings and perform minor medical acts.

Researchers face many flaws in the design of rescue robots. These challenges can be classified in three classes:

- Design and/or conception: Many researchers have proposed single rescue robot [2], [3], [4], [5], [6] that can move in the cluttered environment of disaster sites. These robots use different mechanisms (wheels, legs, tracks and wheeled-legs) to achieve their displacement. However, these robots are not well suited to all disaster environments. The single rescue robot must be carefully chosen in regards to the disaster site. To withstand defaults of single rescue robots, some authors propose to use multiple robots [7], [8], [9]. Each robot can adapt to a precise situation and overcome the defaults of the other ones.
- Positioning: A challenge for rescue robots is to determine their position in a cluttered and unfavorable environment. It is common to use GPS to locate a vehicle or something else. However GPS is not suitable to establish indoor positioning because the GPS signal is highly attenuated and scattered by the rubble from the disaster sites. To localize rescue robots in indoor some authors propose to use the dead reckoning systems of the robot [10], the use of electromagnetic waves for triangulation technique [11], [12] or lasers for scan matching techniques [13]. These proposals offer an accuracy down to a few meters. To monitor the environment efficiently or handle some objects, the positioning system has to be far more accurate. A wise objective would be to obtain a position accuracy of about one or two centimeters, which make it possible to have accurate displacements.
- Human being detection: The main purpose of the rescue robot is to detect human beings. In order to detect human beings, sensors (such as IR sensor, thermal camera, video camera, microphone, ultrasonic sensor, CO2 sensor…) are mounted on rescue robots [14], [15], [16], [17], [18]. To work conveniently, these technologies require to be in close proximity to the victim.

In this paper, we only focus our interest on the two last points. Thanks to the UWB signal characteristics, we propose a method to both achieve efficient human being detection as well as to perform an accurate positioning of the rescue robot.

*Research supported by University of Western Brittany grant LIRAS-UWB.

David Espès is jointly with University of Western Brittany and CNRS UMR 6285 Lab-STICC, 6 Avenue Le Gorgeu, 29238 Brest, France (corresponding author to provide phone: +33298018306; fax: +33298016395; e-mail: David.Espes@univ-brest.fr).

Ana Maria Pistea is with Military Technical Academy, Bd. George Coşbuc, nr. 39-49, Bucharest, Romania (e-mail: ampistea@mta.ro).

Charles Canaff is jointly with University of Western Brittany and CNRS UMR 6285 Lab-STICC, 6 Avenue Le Gorgeu, 29238 Brest, France (e-mail: Charles.Canaff@univ-brest.fr).

Ioana Iordache is with Military Technical Academy, Bd. George Coşbuc, nr. 39-49, Sector 5, Bucharest, Romania (e-mail: ioana.iordache@mta.ro).

Philippe Le Parc is jointly with University of Western Brittany and CNRS UMR 6285 Lab-STICC, 6 Avenue Le Gorgeu, 29238 Brest, France (e-mail: Philippe.Le-Parc@univ-brest.fr).

Emanuel Radoi is jointly with University of Western Brittany and CNRS UMR 6285 Lab-STICC, 6 Avenue Le Gorgeu, 29238 Brest, France (e-mail: Emanuel.Radoi@univ-brest.fr).

UWB systems are currently being developed to help relieve the spectrum drought caused by an explosion of narrowband systems, by offering short-range broadband services, using frequencies already allocated to other applications. Impulse based UWB (IR-UWB) is characterized by the transmission of extremely short duration pulses typically on the order of nanosecond [19], [20].

The use of an extremely large frequency bandwidth makes it possible to measure very precisely indoor distances, e.g. about 5 cm of localization accuracy for a signal bandwidth of 3 GHz. IR-UWB is thus becoming a viable solution for short-range highly accurate localization for difficult environments (e.g. indoor) and critical contexts, such as medical assistance or rescue operations, which are the targeted applications in this paper.

Besides, IR-UWB systems have many other interesting features, such as low complexity and low power consumption, which attract increasing interests in academics and industry. For the application described in this paper, we have been also interested by the low dimensions of the IR-UWB localization system, which makes it appropriate to be mounted on a mobile robot platform.

In this paper, we propose a new method to both detect human body and locate the rescue robots. Due to a precise pattern of the UWB pulse, we can detect human body and locate the rescue robot accurately at the same time.

Unlike other human body detection systems (thermal images, visual images, CO2 detection), our method is not restrained into the close environment of the victim. UWB signals pass through obstacles and can be sensed into a large area. In the same way, ultrason and sound signals may pass through obstacles but with a high attenuation i.e., sounds are detected into a range closer to the victim than with the use of electromagnetic waves. When UWB signals pass through human body, the signal behavior is abruptly modified. Our system focuses on these changes to detect human beings. In disaster situations, it is also important to find dead human bodies. Many relatives or friends of those killed in a disaster ask searching for their dead. Indeed, families need to pay their respects to their dead and wish to have the opportunity to see that person a last time before the remains are returned to the place of the burial. Our technology does not discern between alive and dead human beings so it reports all the human bodies in an area.

To monitor the environment appropriately or perform some minor rescue acts (e.g. first aid, obstacle displacements…) the positioning system has to perform high localization accuracy (about few centimeters). To reach such a precision, we design the UWB pulse to optimize the spectral effectiveness. Hence the symbols are easier to detect and the effects of multipath reflection are reduced. We propose a localization algorithm based on a two steps positioning approach in which certain parameters are extracted from the signals first, and then the position is estimated based on those signal parameters. In the first step, signal parameters, such as the Time of Arrival (ToA) in our case, are estimated using the Dirty Template algorithm. These parameters allow estimating the ranges between the target node and the reference nodes. Then, in the second step, the target node position is estimated based on the signal parameters and range measurements obtained from the first step using the Bancroft algorithm. Our localization algorithm ensures that the precision accuracy is bounded and less than 7 centimeters.

In the first section of this paper, we will concentrate on the system overview, including hardware requirements. In the second section, we will describe the design of the UWB pulse and the interaction between the signal and the environment. In the third section, we present the two steps positioning algorithm and some results. In the last section, we will conclude and give some new research objectives.

## II. System Overview

Two challenges of rescue robots are to offer a precise positioning of the robot and to detect human beings. Our system can achieve these tasks efficiently.

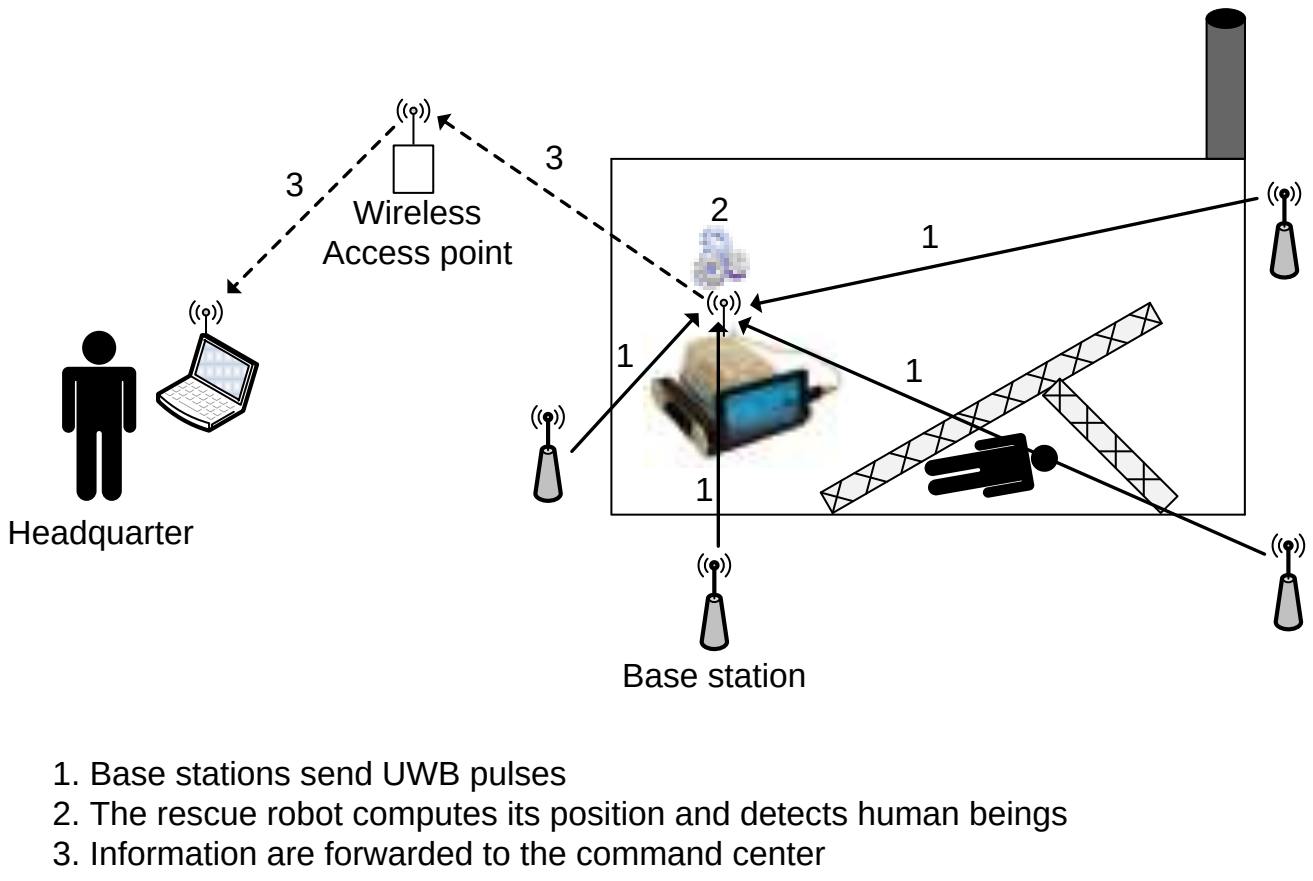

Figure 1. System overview

The system is composed of two types of node, reference nodes and normal nodes. The reference nodes are nodes that know their absolute coordinates. As reference nodes are positioned outside the disaster site, they are equipped with a GPS receiver. The normal nodes do not know their position. They gather information from the reference nodes and estimate their position using multi-lateration technique. As shown in Figure 1, some reference nodes (with known positions) will be distributed around the disaster site. Due to the transmission range of UWB signals in indoor environments, the references nodes cover fully 50 meters. They will send a UWB pulse periodically. A UWB receptor will be added on the rescue robot. The robot will be able to determine its position (3D) after receiving pulses from at least four UWB transmitters, with a 1-2 centimeters accuracy. Due to abrupt changes in the UWB signal, the robot will be also able to determine human bodies.

Due to their radio transceiver, the nodes belong to a same wireless sensor network. Our system complies with the requirements of wireless sensor networks. It is easy to deploy and some nodes can be added or removed. Our system is scalable and a large number of nodes can be operational on the same disaster site. In this paper, results are presented for only 4 reference nodes. Results can be improved while increasing the density of reference nodes.

Due to the flexibility of our system, it can be easily extended to forward location and/or human detection information to the headquarter, see Figure 1.

## III. UWB Pulse Design and Characteristics

### A. *UWB pulse design*

The design of UWB pulse is very important. Thus, the UWB signals have to comply with the FCC masks for the EIRP (Effective Isotropic Radiation Power). This compatibility is necessary in order to ensure sufficiently low levels of power (comparable with the noise) in some spectral bands already occupied by widely used applications and thus avoid introducing disturbances.

At the same time, the allowed spectral domain is to be efficiently occupied, so that maximum energy can be transmitted while remaining compatible with the specifications of the considered FCC mask. The third objective is specifically related to our application, which requires one mobile transmitter and several fixed receivers, closely spaced and operating in the same time slots. Hence, we design several mutually orthogonal UWB pulses in order to be able to distinguish the signals corresponding to different sensors and to minimize their interferences.

Our approach uses a flexible UWB pulse design based on B-spline functions, which can be easily obtained as the output of a simple analog circuit [21].

Thus, $L$ orthogonal UWB pulses $\{\psi_l(t)\}_{l=1..L}$ can be generated as linear combinations of $N_s = L+m-1$ B-spline functions $\varphi_m(t)$, of order $m$:

$$\psi_l(t) = \sum_{k=0}^{Ns-1} c_{l,k} \cdot \varphi_m(t-kT) \ , \qquad l = 1,..,L \tag{1}$$

where $c_{l,k}$ are the weight coefficients obtained from the optimization procedure below.

$$\begin{cases} \min\limits_{c_{l,k}} \left[ -\sum_{l=1}^{L} \xi_l \right] \\ \text{subject to:} \\ \left|\hat{\psi}_l(\nu)\right|^2 \leq S_{FCC}(\nu) \ , \ l = 1..L \\ \sum_{k=0}^{Ns-1} c_{l,k} = 0, \ l = 1..L \\ \int_{-\infty}^{\infty} \psi_l(t) \cdot \psi_p(t) \, dt \ = E_s \delta_{lp} \end{cases} \tag{2}$$

where $\hat{\psi}_l(\nu)$ is the power spectral density of the $l^{th}$ UWB pulse, $S_{FCC}$ is the FCC spectral mask, $E_s$ is the energy of each UWB pulse, $\delta_{lp}$ is the Kronecker symbol and:

$$\xi_l = \frac{\int_0^\infty \left|\hat{\psi}_l(\nu)\right|^2 d\nu}{\int_0^\infty S_{FCC}(\nu) d\nu} \tag{3}$$

is a normalized measure for the effectiveness of the spectral resources use for the $l^{th}$ UWB pulse.

This is a non-linear optimization problem subject to both linear and non-linear constraints, which we solved using genetic algorithms. Figure 2 shows an example of 4 UWB pulses, fulfilling all the required criteria, and generated by our approach for a generic FCC-like spectral mask, with $N_s$ = 30 and $L$ = 4.

### B. *Research framework*

Our experimental UWB platform includes an Agilent 81180A 4.2 GSa/s Arbitrary Waveform Generator (AWG) and E8267D PSG Vector Signal Generator. The first one is able to generate specially designed UWB waveforms with a baseband bandwidth up to 2 GHz, while the second one is used in our experimental setup to shift this large bandwidth in the frequency domain from 3 GHz to 10 GHz. Two sets of UWB antennas, Satimo 050-A and Satimo 300-A, designed to operate up to 1 GHz for the former and between 3 GHz and 10 GHz for the latter, are used for transmitting and receiving the UWB pulses generated with the two generators. The fourth component of our UWB platform is an Agilent DSO90604A Infiniium High Performance Oscilloscope, which has up to 6 GHz of bandwidth and 20 GSa/s on each of 4 analog channels.

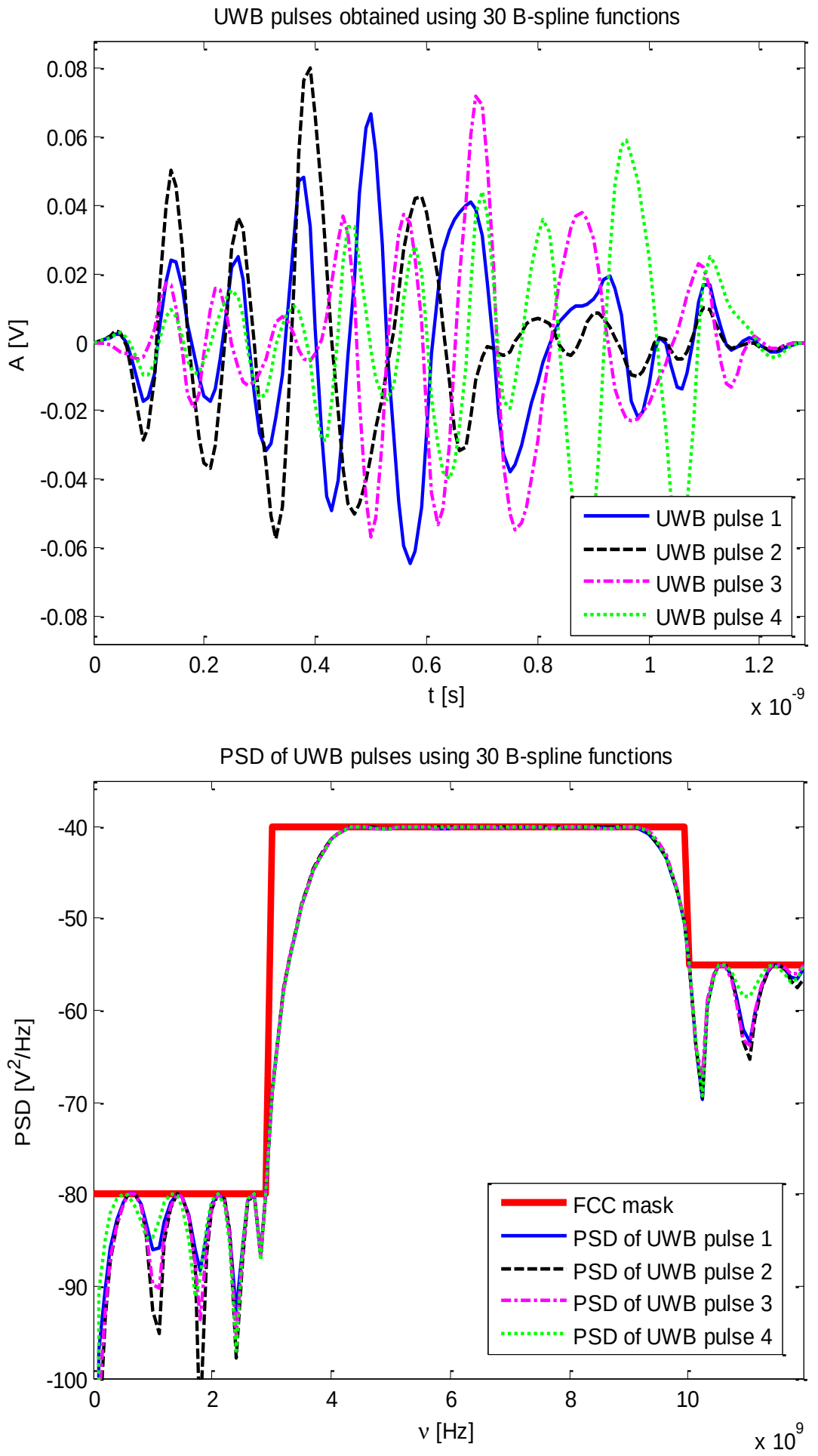

Figure 2. Orthogonal UWB pulses (up) and their PSD compared to the considered spectral mask (down)

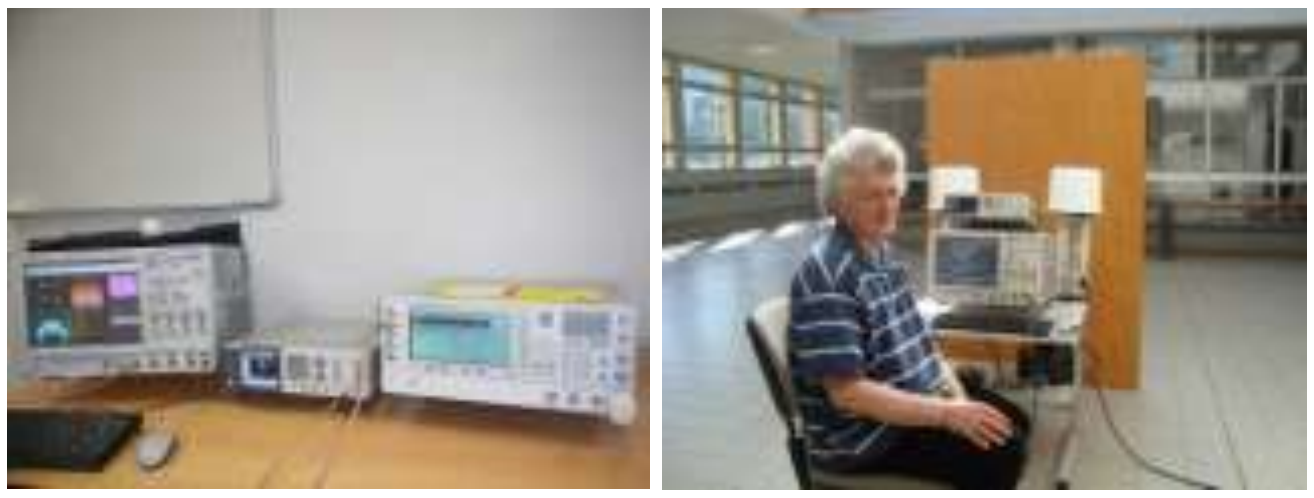

Figure 3. Experimental setup: UWB acquisition system (left) and one of the measurement configurations (right)

### *C. UWB pulse interaction with human body and different environments*

We have used the UWB platform presented above to study the amplitude and phase response of different environments, which are usually impinged and penetrated by UWB signals. The figure 3 provides a zoom, over a bandwidth of 200 Mhz, on attenuation and phase characteristics of these environments. For the sake of presentation convenience the obtained results are presented only over this frequency interval, but they are valid, of course, for the whole UWB signal bandwidth.

Besides the two artificial environments, wood door and brick wall, we have also considered the case of a human body alone and the case when it is hidden behind the door or the wall.

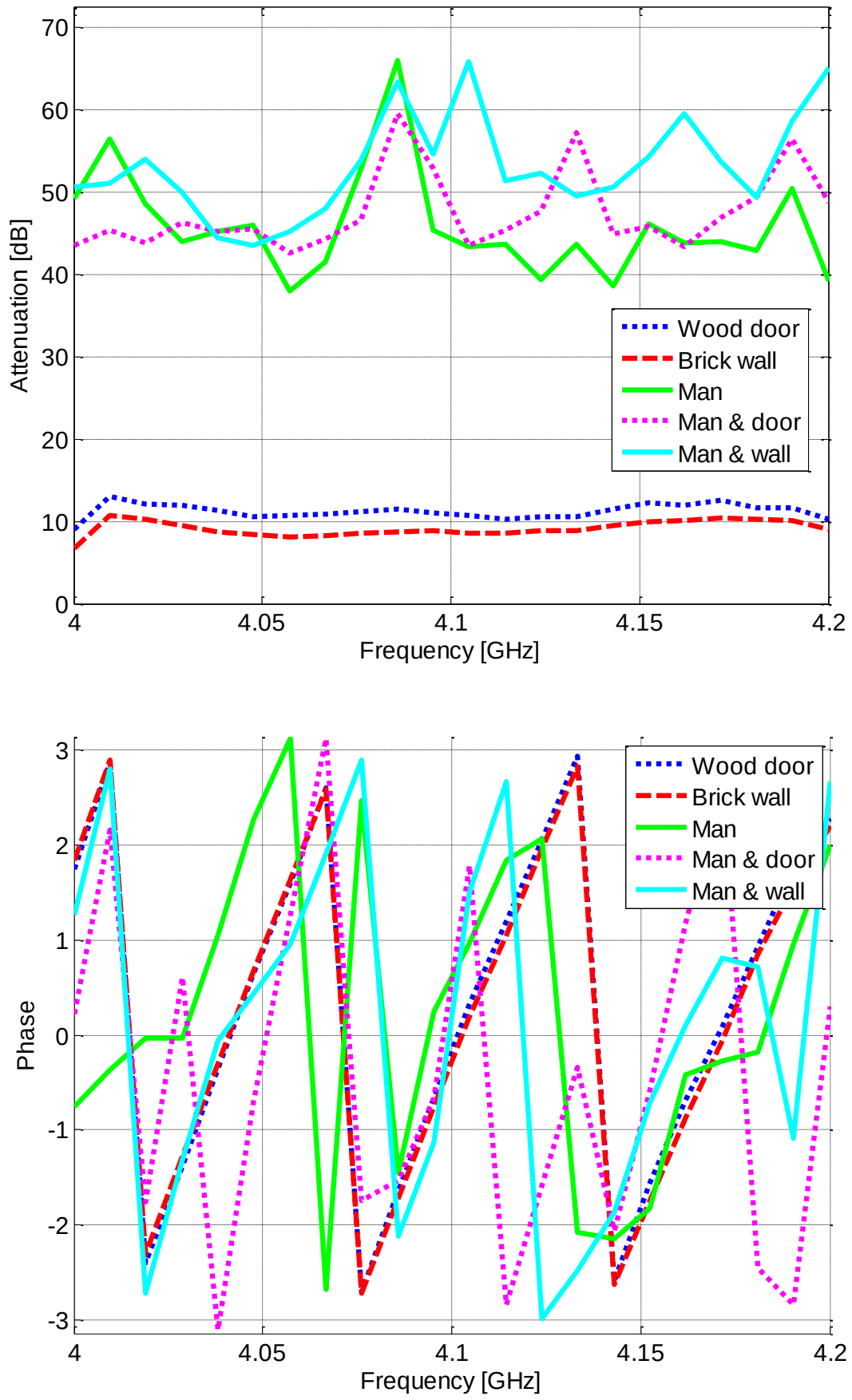

Figure 4. Zoom on attenuation (up) and phase response (down) associated to different test environnements

It is interesting to note that there are two clearly different behaviors, which can be captured by UWB signals and provide useful information for rescue operations. Thus, the two artificial environments have very similar behaviors, characterized by low attenuation level (about 10 dB) and perfectly linear phase response. On the other hand, the presence of a human body modifies dramatically this behavior. Once again, we can note a similar behavior of the 3 corresponding cases (man alone, man behind a door and man behind a wall), characterized by high attenuation level (about 50 dB) and non-linear phase response.

According to these experimental results, we can conclude that the presence of a human body behind some common artificial environments such as wood doors or brick walls, can be detected using its UWB attenuation and phase signatures. Indeed, unless the wood door and brick wall alone, whenever a human body is present the attenuation of the UWB signal is significantly increased and the phase response becomes highly non-linear.

As future work we plan to extend this analysis to the interaction of the UWB signal with some more environments, including some of their combinations of practical relevance.

## IV. Positioning Algorithm

In this section, we present our two steps positioning algorithm. First, the dirty template algorithm determines the time of arrival of the signals. Second, the Bancroft algorithm estimates the position of the target using parameters of the first phase. We present some results to highlight the effectiveness of our solution.

### *A. Dirty Template Algorithm*

In order to perform time-based ranging successfully, the ToA of the received signal should be estimated accurately. The conventional ToA estimation technique is performed by means of matched filtering or correlation operations. Fine time resolution of UWB signals makes accurate identification of the first Multi Path Component (MPC) possible. However, this may not be easy in many scenarios due to non-line-of-sight (NLOS) propagation and a vast number of MPC. Estimation of the ToA could be affected by several errors (multipaths, interference, attenuation, clock drifting, etc.). Therefore, finding a good estimation of the ToA could be a challenging task in the presence of multipath and interference. In this paper, we consider only the LOS case and we use the dirty template algorithm [22] which allows estimating the ToA in the presence of MPC.

The dirty templates algorithm is a low complexity ToA estimator which operates on symbol-rate samples. The basic idea behind this algorithm is as follows. The optimal template signal is not available during ToA estimation. However, the received signal itself can be used as a correlate template, which is noisy ("dirty"). Then, using cross-correlations of the symbol-length portions of the received signal will lead to estimate the ToA. A detailed version of the algorithm may be found in [23].

### *B. Bancroft Algorithm*

In three-dimensional geometry, the trilateration technique uses four reference nodes to calculate the position of the target node. To be localized the target node should locate at the intersection of four spheres centered at each reference position. When the signal received from the reference nodes is noisy, the system is non-linear and cannot be solved. An estimation method has to be used. An estimation of the target position can be solved directly without the process of linearization, thereby not requiring the availability of initial approximate values for the target node position and being non-iterative as a consequence. Bancroft's closed-form solution is global nonlinear least squares method based on the Lorentz metric of hyperbolic space instead of the usual Euclidean metric. The algorithm involves the inversion of a (4,4) matrix and the solution of a scalar equation of second order. Bancroft's method was further discussed and analyzed in [24] and [25].

Solving this second order equation, we get two solutions. Physically, the solution is a real number, otherwise there may be a failure. Suppose now that both solutions are also real, then there will be two solutions. One of these two solutions does not match a real position (e.g. the position is out of bounds or one coordinate is negative...) and needs to be rejected. The other will be the estimation of the real position.

### *C. Results*

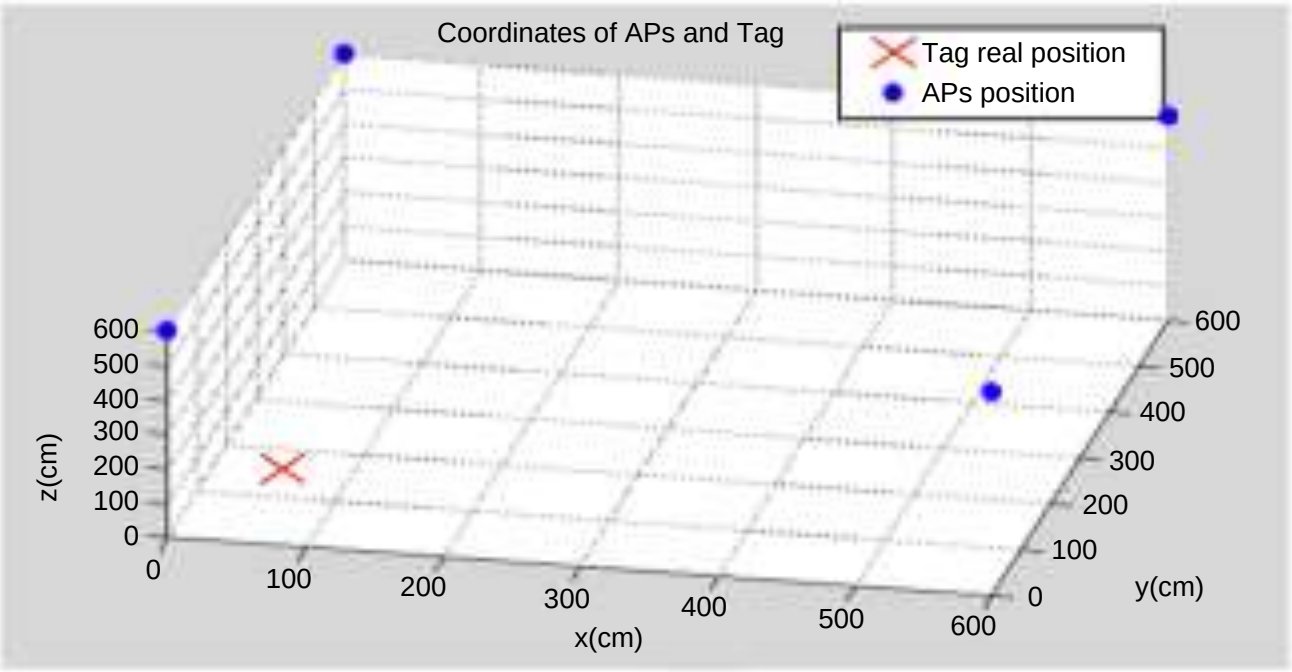

Figure 5. System overview

In this section, we show the results of Matlab simulations. An overview of the system with four reference nodes is shown on Figure 5. In our simulations, four reference nodes are placed in the four upper corners of a room of size 6x6x3 m and the target node is a robot located randomly somewhere on the floor of the room. The reference nodes emit the UWB signals while the target node receives them. To realize the simulation, we use the UWB pulse 1, see Figure 2, with time duration TP=1.28 ns.

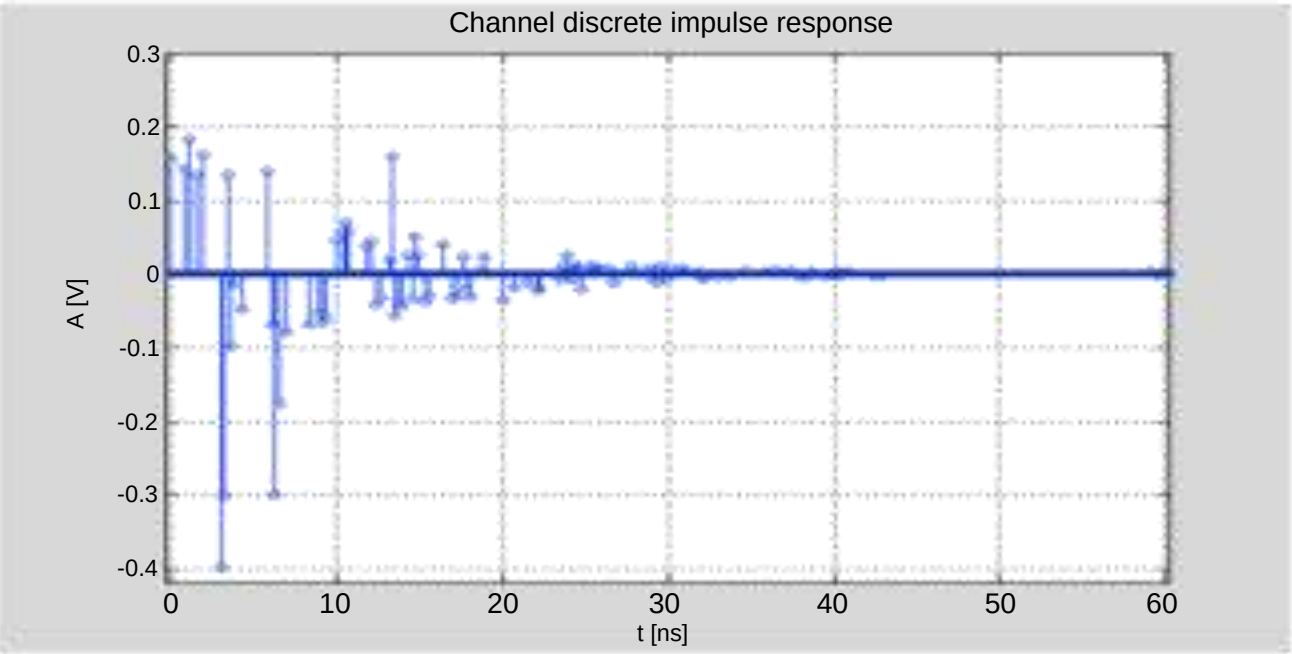

Figure 6. Channel impulse response

The channel model used is a line of sight channel and is characterized by the impulse response given in Figure 6. Each pulse represents a multipath reflection. The multipath reflection duration is higher than the symbol duration. Hence, the interference between symbols is quite important.

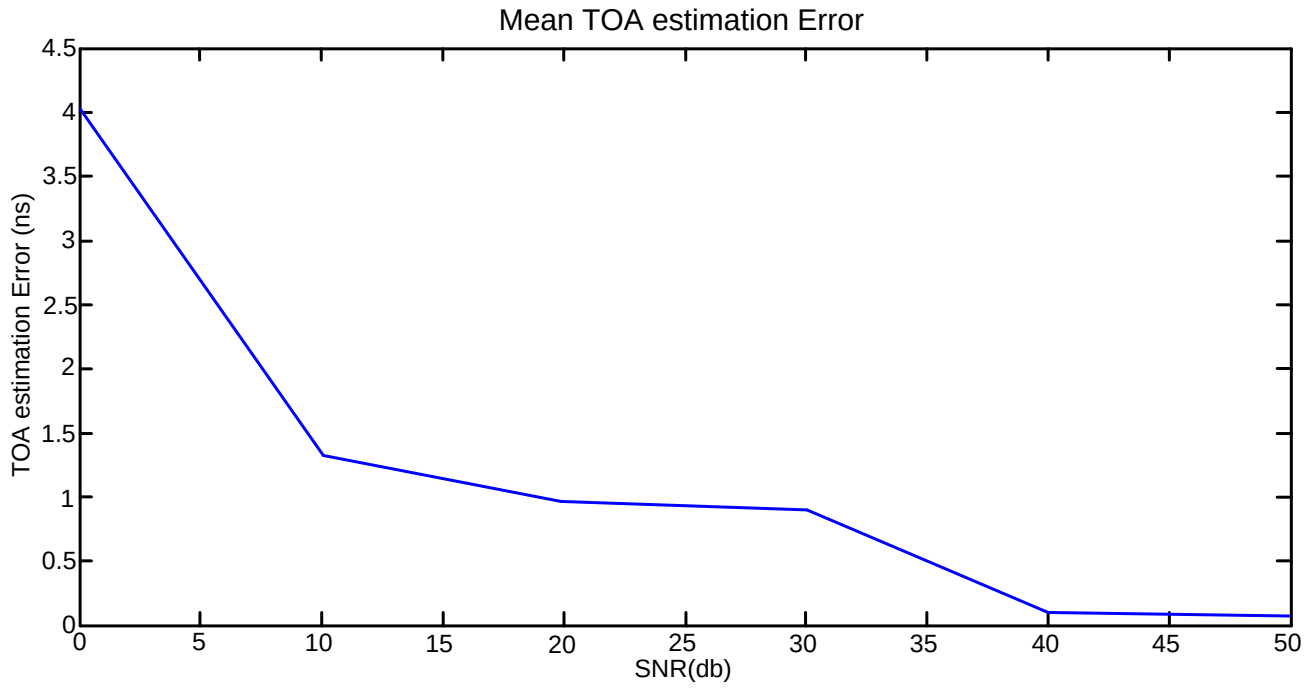

Figure 7. Mean ToA estimation error for reference node 1

As we know, in a two-step positioning algorithm, the positioning accuracy increases as much as the position related parameters in the first step are estimated more precisely. Therefore, in the first step, we analyze the performance of the ToA estimation using the dirty template algorithm. The performance is assessed using the ranging errors of each measurement. Figures 7 and 8 show respectively the ToA estimate and the ranging accuracy concerning the UWB signals of the first reference node for various SNR (Signal to Noise Ratio) values. We show the normalized mean square error which is approximated by the mean of a hundred trials per point. Similar results are obtained for the other reference nodes.

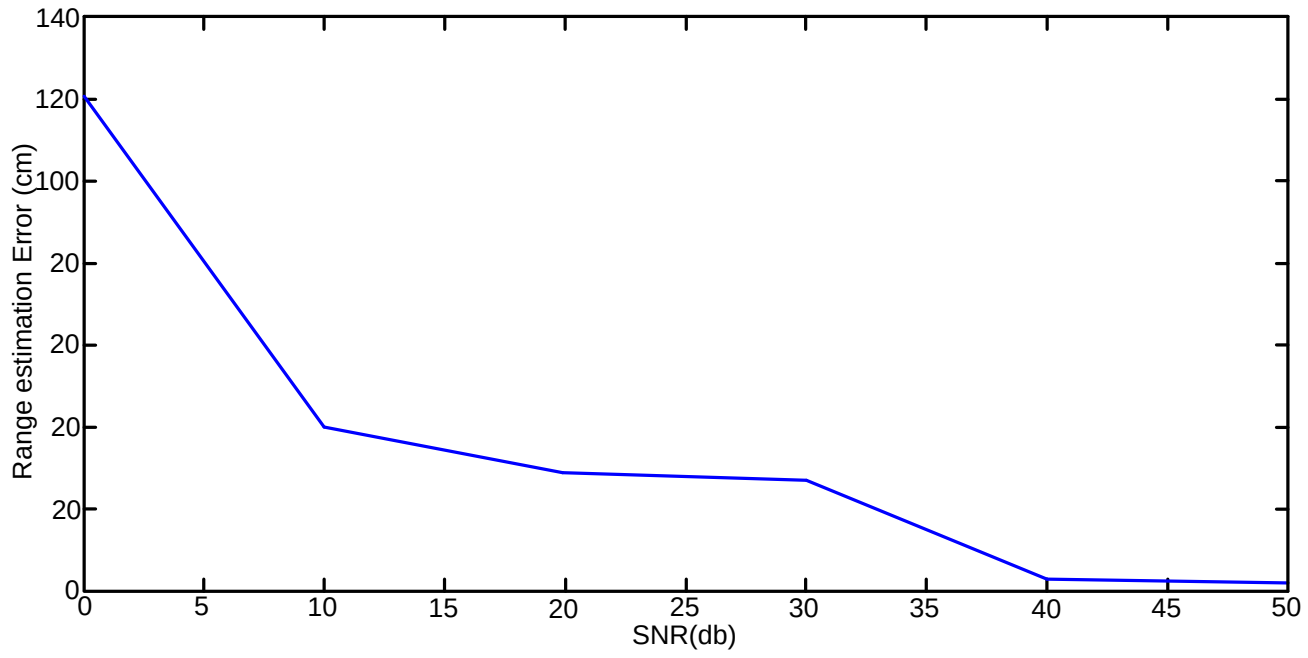

Figure 8. Mean Range estimation error for reference node 1

From Figure 7 we can see that the dirty template allows us to rapidly approach the accuracy limit for high SNR values even in the presence of MPC. Figure 8 shows how the ranging error follows the ToA estimation error, as $\Delta d_i = c\Delta T_i$, and the ranging error decreases rapidly as the SNR increases.

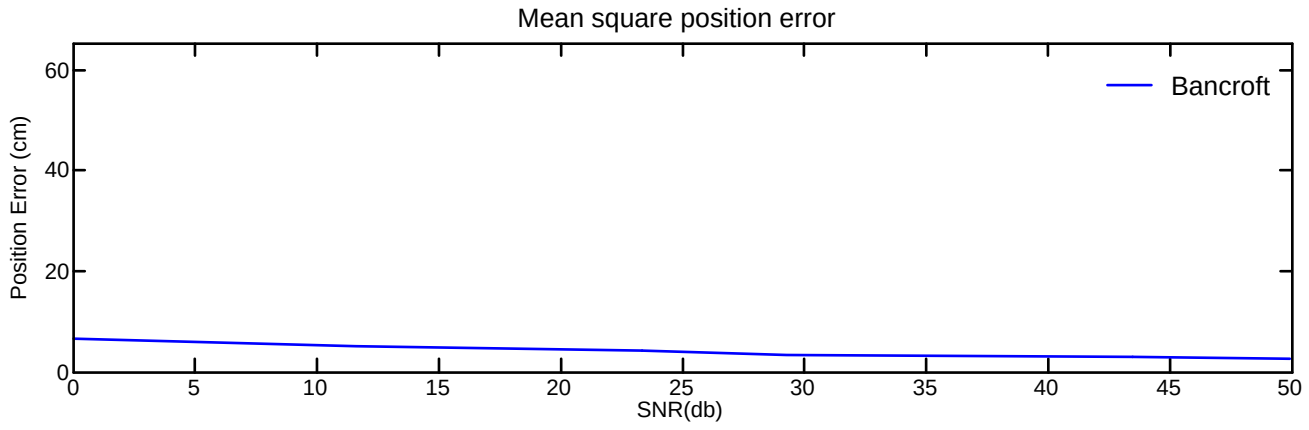

Figure 9. Mean position estimation error

The accuracy of the target node position depends not only on the ToA estimation accuracy but also on the statistical method allowing estimating the position from a certain number of pseudo range measurements. In Figure 9, we analyze the accuracy of the position estimation in terms of the SNR. The performance of the positioning methods is assessed using the position errors regarded as the normalized mean square error of the Euclidean distance between the position estimate and the real position. Whatever the SNR, the precision error does not exceed 7 cm which is quite acceptable. The results show that our solution achieves sufficient accuracy when the SNR is higher than 30 dB. The precision is under 2cm so our location method is well suited for rescue robot positioning systems.

By analyzing Figure 7 and Figure 9, we can conclude some important properties. While the mean square error of the ToA estimate decreases and approaches an asymptotic line (see Figure 7), the positioning error based on Bancroft method decreases independently. As the Bancroft calculation introduces an additional unknown related to the ToA error, this might lead to the suspicion that the ToA delay measurement are somewhat biased.

## V. Conclusion

In cluttered and scattered environments, rescue missions are difficult. Rescue robot is a good solution in order to help rescue teams during human being searches. Some challenges to overcome for rescue robots are to propose new mechanisms to locate the robot and to detect human beings. To ease rescue robot displacements and human being detection, UWB technology is very helpful due to its fine time resolution.

In this paper, we propose a new method to both detect human beings as well as to locate rescue robot accurately. For this purpose we design some UWB pulses on the basis of B-splines. The generated pulses do not only fully comply with the FCC spectral mask but also are highly power efficient in the available spectrum.

Our system can detect human beings due to some changes in the UWB signal shape. Indeed UWB signals change abruptly after passing through a human body. The changes in the UWB attenuation and phase reveal a peculiar signature that can be easily identified. Our system compares the receiving signal with this signature in order to detect human beings.

Our system can also locate rescue robots using the same components. To show the efficiency of our method, we realized intensive simulations on MatLab. The results are quite good in regard to the precision. When the SNR is higher than 30dB, a precision under 2 centimeters is obtained for the target node.

## References

[1] J. Casper and R. Murphy, “Human-robot interactions during the robot-assisted urban search and rescue response at the World Trade Center”, IEEE Trans. On Systems, Man and Cybernetics-part B, vol. 33, pp. 367-385, 2003.

[2] M. Eich, F. Grimminger and F. Kirchner, “Proprioceptive control of a hybrid legged-wheeled robot”, Proc. IEEE Int. Conf. on Robotics and Biometrics, pp. 774-779, Feb. 2009.

[3] Y. Fukuoka and H. Kimura, “Adaptive dynamic walking of a quadruped robot on irregular terrain based on biological concepts”, International Journal of Robotics Research, vol. 22, pp. 187-202, 2003.

[4] M. Ohira, R. Chatterjee and T. Kamegawa, “Development of three-legged modular robots and demonstration of collaborative task execution”, IEEE International Conference on Robotics and Automation, pp. 3895-3900, April 2008.

[5] H. Tsukagoshi, M. Sasaki, A. Kitagawa and T. Tanaka, “Design of a higher jumping rescue robot with the optimized pneumatic drive”, Proc. IEEE Int. Conf. on Robotics and Automation, pp. 1276-1283, April 2005.

[6] M. Suzuki, S. Kitai and S. Hirose, Basic systematic experiments and new type child unit of anchor climber: swarm type wall climbing robot system”, Proc. IEEE Int. Conf. on Robotics and Automation, pp. 3034-3039, May 2008.

[7] T. Kamegawa, K. Saikai and A. Gofuku, “Development of grouped rescue robot platform for information collection in damaged buildings”, The University Electro-Communications, pp. 1642-1647, Aug. 2008.

[8] K. Nagatani, K. Yoshida and K. Kiyokawa, “Development of a networked robotic system for disaster mitigation”, Proc. Field and Service Robotics, pp. 453-462, Aug. 2008.

[9] Y. Yang, G. Xu, X. Wu, H. Feng and Y. Xu, “Parent-child robot system for rescue missions”, IEEE Int. Conf. on Robotics and Biomimetics, pp. 1427-1432, Dec. 2009.

[10] Z. Hu, “Localization system of rescue robot based on multi-sensor fusion”, IEEE Int. Conf. on Computing, Control and Industrial Engineering, Aug. 2011.

[11] S. Miyama, M. Imai and Y. Anzai, “Rescue robot under disaster situation: Position acquisition with omni-directional sensor”, IEEE/RSJ Int. Conf. on Intelligent Robots and Systems, Oct. 2003.

[12] D. P. Stormont and A. Kutiyanawala, “Localization using triangulation in swarms of autonomous rescue robots”, IEEE Int. Work. on Safety, Security and Rescue Robotics, Sept. 2007.

[13] J. Li, J. Bao and Y. Yu, “Study on localization for rescue robots based on NDT scan matching”, IEEE Int. Conf. on Information and Automation, June 2010.

[14] B. Schiele, M. Andriluka, N. Majer, S. Roth and C. Wojek, “Visual people detection: Different models, comparison and discussion”, ICRA Workshop People Detection and Tracking, 2009.

[15] J. Serrano-Cuerda, M. Lopez and A. Fernandez-Caballero, “Robust human detection and tracking in intelligent environments by information fusion of color and infrared video”, Int. Conf. on Intelligent Environments, 2011

[16] A. Himoto, H. Aoyama, O. Fuchiwaki, D. Misaki and T. Sumrall, “Development of micro rescue robot – human detection”, IEEE Int. Conf. on Mechatronics, 2005.

[17] S. Bhatia, H. S. Dhillon and N. Kumar, “Alive human body detection system using an autonomous mobile rescue robot”, IEEE India Conference, Dec 2011.

[18] H. Sun, P. Yang, Z. Liu, L. Zu and Q. Xu, “Microphone array based auditory localization for rescue robot”, Control and Decision Conference, May 2011.

[19] L.Yang and G. B. Giannakis, “Ultra-wideband communications: an idea whose time has come”, IEEE Signal Processing Magazine, vol. 21, no. 6, pp. 26-54, Nov. 2004.

[20] M. Z. Win and R. A. Scholtz, “Impulse radio: How it works”, IEEE Commun. Lett., vol. 2, no. 2, pp. 36–38, Feb. 1998.

[21] M. Matsuo, M. Kamada, H. Habuchi, “Design of UWB pulses based on B-splines”, IEEE International Symposium on Circuits and Systems, vol. 6, pp. 5425-5428, May 2005.

[22] L. Yang and G. B. Giannakis, “Timing ultra-wideband signals with dirty templates”. IEEE Trans. Commun., vol. 53, no. 11, pp. 1952–1963, Nov. 2005.

[23] Z. Sahinoglou, S. Gezici, and I. Guvenc, “Ultra-wideband positioning systems”. Cambridge University Press, 2008.

[24] S. Bancroft, “An algebraic solution of the GPS equations”, IEEE Trans. Aerosp. Electron. Syst., vol. 21, no. 6, pp. 56-59, 1985.

[25] M. Yang and K. H. Chen, “Performance Assessment of a Noniterative Algorithm for Global Positioning System (GPS) Absolute Positioning”, Proc. Natl. Sci. Counc. ROC(A), vol. 25, no. 2, pp. 102-106, 2001